\documentclass[sigconf,natbib=true,anonymous=false]{acmart}

\AtBeginDocument{%
  }

\setcopyright{acmlicensed}
\copyrightyear{2025}
\acmYear{2025}
\acmDOI{XXXXXXX.XXXXXXX}

\usepackage[utf8]{inputenc} 
\usepackage[T1]{fontenc}    
\usepackage{hyperref}       
\usepackage{url}            
\usepackage{booktabs}       
\usepackage{amsfonts}       
\usepackage{nicefrac}       
\usepackage{microtype}      
\usepackage{xcolor}         
\usepackage{graphicx}
\usepackage{subfigure}
\usepackage{multirow}
\usepackage{amsmath, amsthm}
\usepackage{makecell}
\usepackage{tablefootnote}
\usepackage[strings]{underscore}
\usepackage{CJKutf8}
\newcommand{\chinese}[1]{\begin{CJK}{UTF8}{gbsn}#1\end{CJK}}
\usepackage{threeparttable}
\usepackage{fancybox}
\usepackage{mdframed}
\usepackage[title]{appendix}
\usepackage{epigraph}

\begin{document}

\title{MedCT: A Clinical Terminology Graph for Generative AI Applications in Healthcare}

\author{Ye~Chen}
\authornote{Both authors contributed equally to this research.}
\affiliation{%
  \institution{Tiger Research}
  \city{Shanghai}
  \country{China}}
\email{yechen@tigerbot.com}

\author{Dongdong~Huang}
\authornotemark[1]
\affiliation{%
  \institution{Department of Respiratory and Critical Care Medicine}
  \institution{The Fourth Affiliated Hospital of School of Medicine, Zhejiang~University}
  \city{Yiwu, Zhejiang}
  \country{China}}
\email{8016009@zju.edu.cn}

\author{Haoyun~Xu}
\author{Cong~Fu}
\author{Lin~Sheng}
\affiliation{%
  \institution{Tiger Research}
  \city{Shanghai}
  \country{China}}
 \email{{haoyun.xu, cong.fu, lin.sheng}@tigerbot.com}
  
  \author{Qingli~Zhou}
\author{Yuqiang~Shen}
\affiliation{%
\institution{Information Center}
  \institution{The Fourth Affiliated Hospital of School of Medicine, Zhejiang~University}
  \city{Yiwu, Zhejiang}
  \country{China}}
 \email{{zhouql,yuqiangs}@zju.edu.cn}

\author{Kai~Wang}
\authornote{Corresponding author.}
\affiliation{%
 \institution{Department of Respiratory and Critical Care Medicine}
  \institution{The Fourth Affiliated Hospital of School of Medicine, Zhejiang~University}
  \city{Yiwu, Zhejiang}
  \country{China}}
 \email{kaiw@zju.edu.cn}

\renewcommand{\shortauthors}{Chen, et al.}

\begin{abstract}
We introduce the world's first clinical terminology for the Chinese healthcare community, namely MedCT, accompanied by a clinical foundation model MedBERT and an entity linking model MedLink. The MedCT system enables standardized and programmable representation of Chinese clinical data, successively stimulating the development of new medicines, treatment pathways, and better patient outcomes for the populous Chinese community. Moreover, the MedCT knowledge graph provides a principled mechanism to minimize the hallucination problem of large language models (LLMs), therefore achieving significant levels of accuracy and safety in LLM-based clinical applications. By leveraging the LLMs' emergent capabilities of generativeness and expressiveness, we were able to rapidly built a production-quality terminology system and deployed to real-world clinical field within three months, while classical terminologies like SNOMED CT have gone through more than twenty years development. Our experiments show that the MedCT system achieves state-of-the-art (SOTA) performance in semantic matching and entity linking tasks, not only for Chinese but also for English. We also conducted a longitudinal field experiment by applying MedCT and LLMs in a representative spectrum of clinical tasks, including electronic health record (EHR) auto-generation and medical document search for diagnostic decision making. Our study shows a multitude of values of MedCT for clinical workflows and patient outcomes, especially in the new genre of clinical LLM applications. We present our approach in sufficient engineering detail, such that implementing a clinical terminology for other non-English societies should be readily reproducible. With our hope to motivate further research on LLM-based healthcare digitalisation, and at large the wellbeing of humankind, we openly release our terminology, models and algorithms, along with real-world clinical datasets for the development~\footnote{MedCT Github: \url{https://github.com/TigerResearch/MedCT}, MedCT Huggingface: \url{https://huggingface.co/collections/TigerResearch/medct-6744641d6f19b9d70a56f848}}.

\end{abstract}


\received{06 December 2024}
\received[revised]{18 January 2025}
\received[accepted]{18 July 2025}

\maketitle

\section{Introduction}
Standard clinical terminologies, e.g., SNOMED CT, LOINC, ICD, can enable a multitude of values for global healthcare systems. For individual patients and clinicians, terminology or ontology coded electronic health records (EHR) greatly increase the consistency and interoperability of clinical data, and in turn increase the opportunities for real-time decision support for care delivering, retrospective reporting and analytics for research, precision medicine, and management. For populations, standardized clinical information boosts evidence-based healthcare, early identification of emerging health issues, and hence agile response in clinical practices~\cite{LIU2011163,WHO2020,HUSER2012}.	

While clinical terminology, ontology, or knowledge graph has been widely perceived as pivotal for healthcare practice and research, the daunting cost of building and optimization has hindered their wide adoption or effective use. At present, albeit with conventional natural language processing (NLP) techniques, the development of these terminological systems is largely a manual, time-consuming, and error-prone process. SNOMED CT is considered to be the most comprehensive and widely adopted clinical terminology in the world~\cite{Benson2010}. It was established in 1965, has undergone more than twenty years development with over one hundred million dollar investment~\cite{NIH2003}. Yet SNOMED CT still lacks coverage for populous languages such as Chinese, Portuguese, and Arabic. Other deficiencies of SNOMED CT or the similar, are largely inherent in their classical ontological and NLP methodologies. For instance, the dilemma between the presumption of context-free meaning in terminologies and the contextualization nature of human languages~\cite{Linder2009}, the complexity incurred for a robust ontological foundation inevitably sacrificed user friendliness and scalability~\cite{Spackman1998}.

The remarkable advances of large language models (LLMs)~\cite{Brown:2020aa,Touvron:2023aa,Chen:2023ab}, especially the emergent capabilities of semantic understanding, generativeness and interactiveness, have inspired us to explore rapidly building high-quality and reliable terminologies for the healthcare domain. In particular, we address the problem of developing a SNOMED-like~\cite{LIU2011163} terminology for the Chinese clinical domain. With a LLM-based holistic approach, we were able to develop a working version of clinical terminology within three months at a relatively low cost of computing and human labor, i.e., about 100K dollar. Our approach is seamlessly simple, consisting of bootstrapping, truth grounding, entity recognition and linking, and iterative optimization with human in the loop. We report our methodology with implementation details, motivating rapid reproduction for other unattended language and subcultural societies.

As LLMs have been increasingly applied and deployed to real-world healthcare and clinical settings~\cite{Singhal:2023aa,Chen2024aa}, hallucinations or fabricated information, remains one of the prominent challenges, entangled with other constraints including lack of explainability, security and privacy concerns~\cite{Nazi:2024aa}. These limitations are intrinsic in LLMs' probabilistic nature and unsupervised learning paradigm, in which lies their immense power of scaling. However, the safety-critical nature of the healthcare domain requires a more deterministic approach to restraining hallucination, therefore motivating us to explore the other side of the AI world. In particular, we augment LLMs with a model of truth, manifested by a standard terminology, ontology or enriched as a knowledge graph (KG). We took a holistic approach to knowledge augmentation, from raw data standardization, through introducing clinical modality into model training, to graph-augmented or guided generation in inference.

We further deployed the MedCT terminology to a representative spectrum of real world clinical and research applications, harvesting the remarkable capabilities of LLMs, while reining their hallucinations. Our laboratory experiments showed that MedCT terminology and associated models and algorithms achieved SOTA performances in a wide spectrum of clinical tasks, and our field study in real-world setting validated the significant values of the developed terminology, especially in LLM-based applications. To summarize, we believe that we have made the following contributions to the global healthcare system in the AI era, and the rest of the paper is logically structured likewise.

\begin{enumerate}
\item MedCT: the world's first open Chinese clinical terminology at the scale comparable to SNOMED CT.
\item A suite of models and algorithms for readily adoption of the above terminology, namely, MedBERT, a pretrained foundation model, and MedLink, a fine-tuned entity linking model.
\item A holistic approach with implementation details for rapid and cost-efficient development of clinical terminology for other unattended languages.
\item A wide and representative spectrum of real-world clinical applications utilizing the MedCT system, to demonstrate its value propositions and provide a reference framework of truth-augmented LLM applications in the healthcare domain.
\item Finding and observations from the field with regards to the status quo of applying LLMs in real-world clinical setting, e.g., large or small models, LLM or classical NLP techniques, general or domain-specialized models.
\end{enumerate}

\section{Methodology}
We bootstrap our development from SNOMED CT, that is considered to be the most comprehensive and widely adopted clinical terminology, therefore inheriting decades of its achievements. We first applied LLM to contextualize and translate the SNOMED concepts into Chinese, thus forming our initial MedCT terminology. We then collaborated with a tertiary care hospital for truth-grounding the terminology, through annotating real-world EHRs with MedCT while revising the terminology for correction and localization.

At the core of the models and algorithms to utilize MedCT is a clinical foundation model, called MedBERT. We pretrained MedBERT from scratch using a thoughtfully curated clinical dataset, and yielded SOTA performance in semantic understanding. Next, with the MedCT annotated clinical data, we trained MedLink, fine-tuned models for clinical terminology named entity recognition (NER) and linking (NEL), or collectively entity linking (EL). After we deployed MedCT in the field, the learning process is iteratively reinforced, for both the MedCT terminology and entity linking models. Our work is inspired by the SNOMED CT entity linking challenge~\cite{drivendata2024}, and our method largely follows the model-based winning solution SNOBERT~\cite{Kulyabin:2024aa}. We managed to push the boundary further in model performance and multilingual coverage, by leveraging LLMs and well-curated real-world clinical data. We now describe each stage of our method and experimental results.

\subsection{Bootstrapping: contextualized translation}\label{sec-anno}
We first acquired the May 2024 SNOMED CT International Edition release, represented as a directed graph that contains 367,584 vertices (concepts) and 1,215,543 edges (relationships). We focus on three types of hierarchies, namely, body structure, procedure, and clinical finding, by extracting 223,437 concepts (60.8\%) from the whole graph. Table~\ref{tab-medct} illustrates some statistics and examples of our initial MedCT terminology. These three types of terminologies shall account for most clinical significance, while worth the most the entity linking efforts~\cite{drivendata2024}. For example, pharmaceutical and clinical drugs may already be well coded in most healthcare organizations.

\begin{table}
  \caption{MedCT Statistics and Examples}
  \label{tab-medct}
  \scriptsize
  \begin{tabular}{lrl}
    \toprule
    Hierarchies & Concepts (en/zh syn.) & \shortstack{Examples \\ Concept code $\mid$ En syn. $\mid$ Zh syn.}\\
    \midrule
    body structure & 41,964 (72,877) & 35259002 $\mid$ Deltoid muscle $\mid$ \chinese{三角肌}\\
    procedure & 59,119 (99,207) & 50070009 $\mid$ Umbilectomy $\mid$ \chinese{切除脐带} \\
    clinical finding & 122,354 (212,447) & 91936005 $\mid$ Allergy to penicillin $\mid$ \chinese{青霉素过敏} \\
    \cmidrule(r){1-2}
    \textbf{Total} & \textbf{223,437 (384,531)} & \\
  \bottomrule
\end{tabular}
\end{table}

We then employed machine translation to bootstrap a Chinese terminology from English SNOMED CT. Neural machine translation (NMT)~\cite{Stahlberg:2019aa} is the dominant approach today, e.g., Google translate, while directly prompting a generative LLMs can also produce highly competitive translations~\cite{Hendy:2023aa}. Specialized trained NMT models usually take a context-free setting, that is, texts are translated independent of contexts. The context-independent setting is not optimized for short terms such as SNOMED CT, especially for disambiguation (see some examples in Table~\ref{tab-mt}). We leverage the expressiveness and interactiveness of LLMs, to contextualize the terms to be translated into their descriptions. For each concept, we first extracted its fully specified name, synonyms, hierarchy, and relationships from the SNOMED graph. We then composed a collective description from these elements. The prompt to LLM for contextualized translation is: \texttt{\{concept description\}+"\textbackslash n"+In the above context, translate the term \{concept synonym\} into Chinese:}. Intuitively and empirically, the contextual enrichment greatly boosts the LLM's capability to grasp and disambiguate the meaning of terms, as shown by some comparisons in Table~\ref{tab-mt}.

\begin{table}
  \caption{Contextualized Translation via LLM}
  \label{tab-mt}
  \small
  \begin{tabular}{lll}
    \toprule
     Term (hierarchy)& \shortstack{Context-free\\(Google translate)} & \shortstack{Contextualized\\(Tigerbot-3 LLM)} \\
    \midrule
    Meyerson naevus (body) & \chinese{迈耶森很天真} & \chinese{迈耶森氏痣} \\
    Umbilectomy (procedure) & \chinese{脐切除术} & \chinese{切除脐带} \\
    NG feeding (procedure) & \chinese{喂食量} & \chinese{鼻饲喂养} \\
    J3 (finding) & J3 & \chinese{Jaeger第三型} \\
    Emmetropia (finding) & \chinese{正视眼} & \chinese{正常视力} \\
  \bottomrule
\end{tabular}
\end{table}

We used the Tigerbot-3 LLM for contextualized translation, given its superior general-purpose performance and good multilingual and biomedical domain coverage, especially in Chinese~\cite{Chen:2023ab}. We deployed a 100K context window Tigerbot-3-70B-Chat model on a server of $8\times$ NVIDIA RTX4090-24G GPUs. The batch inference for translating 384,531 synonyms took 106 hours (or 848 GPU-hours), with an average length of input being about 500 characters and 180 tokens (mainly description and generation length is small).

\subsection{Truth grounding: linguistic and cultural localization}\label{sec-truth}
With the initial MedCT clinical terminology, we want to further validate the correctness (linguistically and grammatically) and appropriateness (culturally conform to local practices). Meanwhile, we also need an annotated dataset to train the clinical NER and NEL models. We conducted an annotation task to serve both purposes.

In collaboration with a tertiary care hospital in Zhejiang, China, we curated a dataset of clinical notes and EHRs, named as MedCT-clinical-notes. The dataset contains 3,109,181 EHR examples involving 596,680 patients, collected from 23 clinical departments over the first quarter of year 2024. The data was de-identified and organized into a format similar to MIMIC-IV free-text clinical notes~\cite{Johnson:2023}, see Appendix~\ref{apdx-ehr} for some example snippets. Notably, our clinical data is significantly more comprehensive than MIMIC-IV notes, from patient's biography and social history information, to daily ward admission records, and even atrial fibrillation follow-ups if applicable. We curated this clinical data for multifold purposes, not only the entity linking task but also downstream applications. With these desiderata and the auto-regressiveness of generative LLM, we structured the clinical notes along the patient dimension, following a temporal or causal order. This way, a patient's clinical data can be analogous to a text document, one document per patient. This data structure is so designed for LLM-based clinical applications, including EHR auto generation, treatment pathway simulation for assistive diagnosis and treatment. We then randomly sampled 1,000 patients' clinical data or ``document'' for annotation.

We formed a group of 14 doctors with at least five years clinical experiences. These physician were instructed to perform two tasks of annotation at the same time as follows.
\begin{enumerate}
\item Annotate and link spans of text in clinical notes with specific concepts in the MedCT clinical terminology.
\item Comment linguistic and translation mistakes and cultural inappropriateness based on their practices. 
\end{enumerate} 
Both the Chinese MedCT and original English SNOMED CT terminologies were given to the doctors for cross validation. Two sessions of in-depth and hands-on training were given to both the management and physicians, and the annotation tasks were facilitated by the open-source annotation tool doccano~\cite{Nakayama2018}. For corrections and additions suggested from the comments, if more than half of the annotators suggests the same and the director doctors of the relevant departments approve, the amendments will be merged into the next release of MedCT.

We emphasized the quality and even serendipities during the annotation process, since essentially the entity linking model to be trained is learning and encoding the domain expertises from the annotator clinicians' minds. The annotators were encouraged to discuss with their non-annotator fellow clinicians, and to consult with official literature and thesaurus, should there be any questions. For the sampled 1,000 patients' EHRs or 7,437 clinical notes (notes can be considered as different sections in a patient's document, e.g, diagnosis and treatment plan, brief hospital course, and discharge instructions). The group of annotators exercised at a carefully balanced speed of about 200 notes per day, so it took more than one month to annotate the dataset. As a result, the annotated dataset contains 61,660 entity mentions or about 8 annotations per note. Table~\ref{tab-note-ann} illustrates some examples of clinical notes with annotated MedCT concepts. More intriguingly, some anecdotical findings during annotation from the field are summarized in Table~\ref{tab-ann-finds}.

\begin{table}
  \begin{threeparttable}
  \caption{Clinical Note Snippets with Concept Annotations}
  \label{tab-note-ann}
  \tiny
  \begin{tabular}{ll}
    \toprule
    Notes  & Concept ID (hier.), name \& syn. \\
    \midrule
   \multirow{3}{*}{\ldots\chinese{右肾小结石。}\colorbox{green!50}{\chinese{右肾上腺}}\tnote{1}\ \ \chinese{结节，建议进一步检查。}} & 29392005 (body) \\
   & Right adrenal gland \\
   & \chinese{右肾上腺} \\
   \midrule
    \multirow{3}{*}{\ldots\colorbox{blue!30}{\chinese{支气管扩张试验}}\chinese{：吸入沙丁胺醇400}ug\ldots} & 415299008 (procedure) \\
    & Reversibility trial by bronchodilator \\
    & \chinese{支气管扩张剂可逆性试验} \\
    \midrule
   \multirow{3}{*}{\ldots\colorbox{red!30}{\chinese{前列腺钙化灶}}\chinese{。左侧精囊腺及两侧输精管钙化。}} & 1259248003 (finding) \\
   & Calcinosis of prostate \\
   & \chinese{前列腺钙化} \\
   \midrule
   \multirow{3}{*}{\ldots\chinese{左中上腹肠系膜}\colorbox{red!30}{\chinese{脂膜炎}}\chinese{考虑。}} & 22125009 (finding) \\
   & Panniculitis \\
   & \chinese{脂膜炎}\\
   \midrule
   \multirow{3}{*}{\ldots\chinese{附见：}\colorbox{red!30}{\chinese{两侧胸腔少量积液}}\chinese{伴邻近肺不张。}} & 425802001 (finding) \\
   & Bilateral pleural effusion \\
   & \chinese{双侧胸腔积液} \\
   \midrule
   \multirow{3}{*}{\ldots\chinese{自病来，神志清，精神一般，}\colorbox{red!30}{\chinese{胃纳差}}\chinese{，睡眠差}\ldots} & 64379006 (finding) \\
   & Decrease in appetite \\
   & \chinese{没胃口，食欲下降} \\
  \bottomrule
\end{tabular}
\begin{tablenotes}
\item[1] We color-coded concept hierarchy as: \textcolor{green!100}{green-body}, \textcolor{blue!100}{blue-procedure}, and \textcolor{red!100}{red-finding}.
\end{tablenotes}
\end{threeparttable}
\end{table}

\begin{table}
  \caption{Findings from Annotation}
  \label{tab-ann-finds}
  \tiny
  \begin{tabular}{llc}
    \toprule
    Findings & Terms in notes  & Concept ID \& name \\
    \midrule
   Not native Chinese & \chinese{排便时出现}\colorbox{red!30}{\chinese{黑便}} & \shortstack{35064005 \\ \chinese{深色大便}}\\
   \midrule
  \multirow{3}{*}{Synonymous mentions} & \colorbox{red!30}{\chinese{纳差 / 胃纳差}} & \shortstack{64379006 \\ \chinese{没胃口；食欲下降}} \\
   & \colorbox{red!30}{\chinese{腹软 / 腹平软}}& \shortstack{249543005 \\ \chinese{腹部柔软}}  \\
   & \colorbox{blue!30}{\chinese{支气管扩张 / 舒张试验}}& \shortstack{415299008 \\ \chinese{支气管扩张剂可逆性试验}} \\
   \midrule
    \multirow{3}{*}{Redundant concepts} & \colorbox{red!30}{\chinese{流涕}}& \shortstack{275280004 / 64531003 \\ Sniffles / Nasal discharge} \\
   & \colorbox{red!30}{\chinese{甲状腺癌}} & \shortstack{363478007 / 448216007 \\ Thyroid cancer / Carcinoma of thyroid} \\
   & \colorbox{red!30}{\chinese{出院}} & \shortstack{183665006 / 183667003 \\ Discharged from hospital / inpatient care} \\
   \midrule
    \multirow{3}{*}{Part-of concepts} & \colorbox{blue!30}{\chinese{腹部彩超}}& \shortstack{438416007 \\ Ultrason. of abdomen and urinary} \\
   & \colorbox{blue!30}{\chinese{泌尿系彩超}} & \shortstack{438416007 \\ \chinese{腹部及泌尿系统超声检查}} \\
  \bottomrule
\end{tabular}
\end{table}

\subsection{MedBERT: a clinical foundation model}\label{sec-medbert}
Of the central importance for most clinical NLP tasks is a foundation model to encode broad and basic semantics in the domain. Previous work shows that for domains with copious amounts of unlabeled texts, pretraining language models from scratch yielded substantial gains over continual pretraining from general-domain models~\cite{Gu:2020aa}. Biomedicine is one of such high-resourced domains. Specifically, we pretrain a BERT model from scratch using a biomedical dataset curated with the following design considerations, with statistics and sources outlined in Table~\ref{tab-medbert-data}.

\begin{enumerate}
\item A large corpora of biomedical literature and publications with comprehensive and timely coverage of the domain, e.g., the PubMed Central (PMC) repository~\cite{Beck2010}.
\item Data from the field and directly relevant to downstream tasks, e.g., clinical guidelines~\cite{cco2024,cdc2024,cochrane2024,nice2024,who2024} and real-world clinical notes MIMIC-IV~\cite{Johnson:2023}.
\item Clinical terminologies and their contexts, e.g., SNOMED CT and MedCT terms and descriptions.
\item Multilingual coverage, i.e., English and Chinese. 
\end{enumerate}

%

\begin{table}
  \caption{MedBERT Training Data}
  \label{tab-medbert-data}
  \footnotesize
  \begin{tabular}{llrr}
    \toprule
    Source  & Dataset (lang) & Examples & Disk size \\
    \midrule
    \multirow{7}{*}{Publications} & PMC abstracts (en)~\cite{Beck2010} & 24,732,786 & 26G \\
    & PMC full-texts (en)~\cite{Beck2010}  & 3,775,772 & 109G \\
    & PMC patients (en)~\cite{Zhao:2023aa} & 167,034 & 444M \\
    & PubMedQA contexts (en)~\cite{Jing2019} & 211,269 & 280M \\
    & Open medical books (en)~\cite{walther20} & 13,000 & 11G \\
    & Chinese literature (zh)~\cite{zju2024} & 27,704 & 14G \\
    & Trad. Chinese medicine books (zh)~\cite{guo1988} & 17 & 13M \\
    \midrule
    \multirow{2}{*}{Guidelines} & Clinical guidelines (en)~\cite{cco2024,cdc2024} & 11,184 & 527M \\
    & Clinical guidelines (zh)~\cite{cai2024,dai2021} & 4,364 & 643M \\
    \midrule
    \multirow{2}{*}{Clinical notes} & MIMIC-IV v2.2 clinical notes (en)~\cite{Johnson:2023} & 2,653,148 & 5.8G \\
    & Chinese EHR and clinical notes (zh) & 3,109,181 & 904M \\
    \midrule
    Terminology & SNOMED and MedCT (en)~\cite{LIU2011163} & 723,552 & 23M \\
    \midrule
    \textbf{Total} & --- & \textbf{35,429,011} &  \textbf{168G} \\
  \bottomrule
\end{tabular}
\end{table}

\begin{table}
  \caption{MedBERT Evaluation}
  \label{tab-medbert-eval}
  \small
  \begin{tabular}{llr}
    \toprule
    Type & Model & Accuracy \\
    \midrule
    \multirow{5}{*}{Biomed} & BiomedBERT-base-fulltext~\cite{Gu:2020aa} & 0.5633 \\
    & BiomedBERT-large-abstract  & 0.5100 \\
    & BiomedBERT-base-abstract & 0.4209 \\
    & SciBERT~\cite{Beltagy:2019aa} & 0.5819 \\
    &\textbf{MedBERT} & \textbf{0.8344} \\
    \midrule
    \multirow{4}{*}{General} & BERT-base-multilingual~\cite{Devlin:2018aa} & 0.5333 \\
    & BERT-base-Chinese & 0.5582 \\
    & BERT-large & 0.3199 \\
    & BERT-base & 0.3440 \\
  \bottomrule
\end{tabular}
\end{table}


We compared the prediction accuracy of the fill-mask task between our MedBERT and other SOTA biomedical and general-domain models, as the results exhibited in Table~\ref{tab-medbert-eval}. First, we verified that domain-specific training has advantages, as biomedical models outperform general-domain BERT models by about twenty percentage points. Second, multilingual expansion is critical. Although the evaluation dataset only has less than 10\% Chinese data, the multilingual and Chinese BERT surpass the English-only models by a large margin. Furthermore, the scale and quality of the training data tends to yield better model performance, as seen that BiomedBERT trained with PMC full text wins those with PubMed abstracts only. 

Nevertheless, our model MedBERT achieves substantial gains over both the biomedical SOTA and general-domain BERT models. Most popular biomedical models were primarily trained on scientific papers in the domain, e.g., BiomedBERT used PubMed~\cite{Beck2010} and SciBERT was trained on Semantic Scholar papers~\cite{Ammar:2018aa}. Our MedBERT training corpus mixed in about 20\% data from the field that is directly concerning downstream clinical tasks, i.e., clinical guidelines and protocols, real clinical notes, and terminologies. Healthcare is such a high-resourced domain that data shift may appear as domain shift, conjecturally explaining the performance gain of MedBERT over other popular models.

\subsection{MedLink: clinical entity recognition and linking}\label{sec-medlink}
We implemented a two-stage approach to recognizing clinical entities from free-text notes and linking the entities to the built MedCT concepts, as follows.
\begin{enumerate}
\item First stage: A NER segmentation task to detect spans of texts as clinical entity mentions.
\item Second stage: A NEL ranking task to predict the MedCT concepts for the recognized entities from the first stage.
\end{enumerate}

For the first stage NER task, we fine-tuned a token classification model from the MedBERT foundation model, as described in Section~\ref{sec-medbert}. We classify each token into four classes: \texttt{\{finding, procedure, body, none\}}, using the BIO format~\cite{ramshaw-marcus-1995-text}, therefore a token tagging task with seven labels: \texttt{\{O, B-find, I-find, B-proc, I-proc, B-body, I-body\}}. We trained the NER model on the annotated clinical note data described in Section~\ref{sec-anno}. The annotated data contains 1,000 patients, 7,437 clinical notes, and 61,660 entity mention annotations. Our annotated dataset is ten times larger than that of the SNOMED CT challenge~\cite{drivendata2024}, in both size and annotation examples. Without specially instructed, our annotated classes follows a similar distribution ($\text{find}:\text{proc}:\text{body}=0.59:0.18:0.23$) as SNOMED CT challenge annotation data ($\text{find}:\text{proc}:\text{body}=0.54:0.34:0.12$), except for the class \texttt{procedure}. Further investigation with clinicians reveals that the data discrepancy largely comes from the healthcare gap between China and America, which is in turn influenced by various factors such as cultural preferences, medical standards, and insurance coverage. One reported example is the popularity difference in C-Section~\cite{Gong:2014}. These discrepancies in healthcare practices underscore the value proposition from localization of clinical terminologies such as the MedCT in our endeavor.

We first split the data into four folds with random shuffling, used fold 1 to 3 for training and held out fold 0 for validation. The tokenized data is chunked into a max sequence length of 512, and repeated 10 times for each step. We fine-tuned for 200 epochs over the training data with a batch size of 8, a initial learning rate of 5e-5 and linear decaying schedule.

At the second stage NEL task, we need to link segmented entity mentions or text spans to concepts in the MedCT ontology. This is a semantic matching task, which we therefore simply formulate it as a ranking problem in the embedding space. We chose the SapBERT models for embedding~\cite{Liu:2020aa,Liu:2021aa}, which are fine-tuned BERT models specially for aligning biomedical synonyms by leveraging the UMLS dataset~\cite{nih:2024}. Specifically, we used \texttt{SapBERT} for English tasks, and its cross-lingual extension \texttt{SapBERT-all-lang} for multilingual and Chinese applications. For each concept from the MedCT terminology, we first embed each synonyms into a latent space of length 512, and then average over synonym embeddings as the concept representation. In inference time, we calculate the embeddings for the recognized entity mentions, and rank to predict their corresponding concepts from the MedCT ontology embedding database using cosine similarity.

We measure the performance of trained models with character-level concept-averaged intersection-over-union (IoU) defined as follows,
\begin{align}
	&\text{IoU}_\text{concept}=\frac{P^\text{char}_\text{concept} \cap G^\text{char}_\text{concept}}{P^\text{char}_\text{concept} \cup G^\text{char}_\text{concept}}\\
	&\text{IoU}_\text{all}=\frac{\sum_{\text{concept} \in P \cup G}{\text{IoU}_\text{concept}}}{N_{\text{concept} \in P \cup G}}
\end{align}
where $P$ and $G$ denote predicted and ground-truth character-concept assignment, respectively. For each experiment, we keep the checkpoint with the highest IoU score as candidate model. We conducted two sets of experiments, one on English MIMIC-IV data (same as SNOMED CT challenge) and the other on Chinese clinical notes (as used for MedCT). Table~\ref{tab-medlink-eval} exhibits the experimental results. Our MedLink model achieves SOTA performance in both English and Chinese clinical NER and NEL tasks. We deem that a stronger multilingual foundation model MedBERT and copious annotated real-world clinical training data largely contribute into the gain.

\begin{table}
  \caption{MedLink Evaluation}
  \label{tab-medlink-eval}
  \footnotesize
  \begin{tabular}{llrrr}
    \toprule
    Type & Base model & \shortstack{English NEL \\ (IoU on MIMIC)} & \shortstack{Chinese NEL \\ (IoU on MedCT)} \\
    \midrule
    \multirow{5}{*}{Biomed} & BiomedBERT-base-fulltext~\cite{Gu:2020aa} & 0.4797 & 0.0091 \\
    & BiomedBERT-large-abstract  & 0.4952 & 0.0005 \\
    & BiomedBERT-base-abstract & 0.4976 & 0.0003 \\
    & SciBERT~\cite{Beltagy:2019aa} & 0.4993 & 0.0026 \\
    &\textbf{MedBERT} & \textbf{0.5065} & \textbf{0.3012} \\
    \midrule
    \multirow{4}{*}{General} & BERT-base-multilingual~\cite{Devlin:2018aa} & 0.4717 & 0.1006 \\
    & BERT-base-Chinese & 0.4508 & 0.1516 \\
    & BERT-large & 0.4868 & 0.0007 \\
    & BERT-base & 0.4774 & 0.0002 \\
  \bottomrule
\end{tabular}
\end{table}


\subsection{Iterative reinforcement: human in the loop}
As we deployed our MedCT system in the field to a tertiary care hospital in Zhejiang, China. We took a reinforcement learning approach to iteratively optimize the terminology, for both coverage and precision. A sample of 1,000 MedCT tagged clinical notes was reviewed by physicians on a monthly basis, mistakenly tagged examples were corrected and then fed into the NER and entity linking models for continual fine-tuning, using a rejection sampling mechanism. If a certain amount of wrongly tagged examples was attributed to the lacking in the terminology, clinicians would amend those missing concepts into the terminology.

\section{Experiments and Applications}
\subsection{Large or small models}

Other than our approach of specialized trained BERT models, directly using generalized pretrained LLMs for NER and NEL is intriguingly appealing. MedCT is a bidirectional transformers model, while LLMs are autoregressive models such as GPT. Like most BERT models, MedCT is relatively small in model size, with 438M parameters in our release. LLMs on the other hand, typically have several hundred billion parameters, like GPT or Llama. Large models have demonstrated substantial improvements on a wide spectrum of NLP tasks, while remain highly general paradigms for training, inference and deployment. This generality is particularly appealing for cost efficient application development, which usually amount to few-shot prompt engineering or lightweight task-specific fine-tuning.

However, training and inferring general-purpose LLMs is dauntingly expensive today. For example, it took 7.0M NVIDIA H100 GPU hours or about 13.7M US dollars to train Llama-3.1-70B~\cite{llama3:2024}. Besides development and recurring computational cost, It remains unclear whether general autoregressive language models can accurately capture complex domain structures. Small yet task-specific designed and trained models may lack the versatility hence requires one-time development effort. Nevertheless, its potential for deep specialization, agile experimentation, and low cost for deployment and use, presents benefits for critical and infrastructural tasks.

In this experiment, we compared the two methodologies in the medical NER and NEL applications: small specialized models versus large generalized models (LLMs). For the LLM approach, we first prompt the OpenAI GPT-4o~\cite{gpt4o:2024} for the NER task. After a mild prompt engineering, we use a simple few-shot prompt as showed in Figure~\ref{fig-fewshot-prompt}.

\begin{figure}
\centering
\begin{mdframed}
\small
\begin{verbatim}
# Few-shot prompt for NER:
{clinical note}

From the above clinical notes, extract all medical entity 
mentions as span of texts, and categorize them into 
three types: body, procedure, and finding. Output as 
JSON format as the following examples:
{"mention": "stomach", "type": "body"},
{"mention": "nasojejunal feedings", "type": "procedure"},
{"mention": "multisystem organ failure", 
"type": "finding"}.
Please only output the JSON result.
\end{verbatim}
\end{mdframed}
\caption{A few-shot prompt for LLM-based NER}
\label{fig-fewshot-prompt}
\end{figure}

The LLM applies its general world knowledge and the few-shot examples provided to perform the NER task. With the presumably state-of-the-art LLM GPT-4o, the result seems impressive, as one snippet showed in Figure~\ref{fig-llm-ner-snippet}.

\begin{figure}
\begin{mdframed}
\small
\begin{verbatim}
# clinical note:
note_id,text
10575317-DS-6,"
Name:  ___                    Unit No:   ___
...
ROS: (+) Back/shoulder pain with emesis, (+) epigastric pain 
with emesis, (+) chills, (+) 8 lb weight loss. Denies fever, 
myalgias, diarrhea, SOB, dizziness, rhinorrhea, cough. 
No sick contacts. Seasonal flu shot 1 month ago. No H1N1.
...

# GPT-4o output:
10575317-DS-6,"```json
{"mention": "emesis", "type": "finding"},
{"mention": "chills", "type": "finding"},
{"mention": "epigastric pain", "type": "finding"},
{"mention": "back pain", "type": "finding"},
...
```"
\end{verbatim}
\end{mdframed}
\caption{A snippet from LLM-based NER}
\label{fig-llm-ner-snippet}
\end{figure}

In the second stage NEL task, we apply the same approach as in MedLink (as described in Section~\ref{sec-medlink}). The LLM tagged entity mentions were linked to the MedCT concepts using embedding similarity match. In the training stage for both the small and large model approaches, we count the most frequent concept for each entity mention from the annotated training dataset. This gives a static dictionary (mention $\rightarrow$ concept) to simply lookup text spans for concepts by phrase matching. But this static dictionary still requires a specially annotated training dataset. In this experiment, we will also conduct an ablation study to analyze the value of the static dictionary lookup.

We ran the LLM-based NER and NEL experiments in both English and Chinese settings. For the English NEL task, we used a left-out validation dataset of 51 MIMIC-IV clinical notes (one fourth of the SNOMED challenge data). For the Chinese experiment, we also used the validation dataset of 1,860 clinical note snippets from the MedCT training data built in-house (as elaborated in Section~\ref{sec-truth}). The experimental results are shown in Table~\ref{tab-medlink-llm}.

Let us first consider the most generalized LLM approach that does not rely on any specialized data or static dictionary lookup. Under this setting, GPT-4o only yields 0.11 IoU on English data and 0.17 IoU on Chinese data, substantially inferior to our MedLink small model approach (0.51 IoU for English and 0.30 IoU for Chinese). Although the LLM results are visually appealing as illustrated above, its numerical measurement of performance is suboptimal. By adding static dictionary lookup, the English IoU improved to 0.35 and Chinese metric barely changes to 0.18, still a considerable gap from MedLink. Noticeably, simple lookup from training set dictionary accounts for more than half of the performance with LLM approach, in English test. In Chinese data, the static lookup barely contributes to performance metrics, primarily because of less overlapping of entity mentions between training and validation sets. Moreover, LLM approach with GPT-4o incurs considerably more inference time than small model approach with MedCT, more than 10 times for English and 30 times for Chinese tests.

We also experimented with Llama-3.1-70B, one of prominent open-source LLMs. Somewhat unexpectedly, Llama-3.1-70B achieved comparable accuracy performance as the proprietary GPT-4o in our medical NER and NEL tasks, in both English and Chinese evaluations. Open-source alternatives are attractive since they are white-box and cost-efficient solutions. Our empirical results show that, for domain-specific tasks with some degree of complexity, the performance gap between close and open models may become negligible. All evaluations were ran on mainstream compute resources. The LLM approach with GPT-4o used OpenAI API, Llama-3.1-70B was deployed on a local machine with 8$\times$ NVIDIA A800-80G GPUs, while MedLink ran on one NVIDIA A800-80G GPU. Overall, besides the unsatisfactory accuracy performance, both the slower response time and the expensive inference hardware makes even the SOTA LLMs less appealing for broad adoption on fundamental tasks such as medical NER and NEL. 


\begin{table}
\begin{threeparttable}
  \caption{MedLink vs. LLM approach}
  \label{tab-medlink-llm}
  \footnotesize
  \begin{tabular}{llrrrr}
    \toprule
    \multirow{2}{*}{Model} & \multirow{2}{*}{Static} & \multicolumn{2}{c}{\shortstack{English NEL\\ (51 MIMIC notes)}} & \multicolumn{2}{c}{\shortstack{Chinese NEL\\(1860 MedCT notes)}} \\
   & & IoU & Time & IoU & Time \\
    \midrule
     \multirow{2}{*}{\textbf{MedLink}} & w/ static & \textbf{0.5065} & 1m40s &\textbf{0.30117} & 4m15s \\
     & w/o static & 0.4320 & 1m23s & 0.30118 & 4m4s \\
     \midrule
    \multirow{2}{*}{GPT-4o\tnote{1} }& w/ static & 0.3493 & 13m57s & 0.1798 & 116m52s \\
    & w/o static & 0.1146 & 13m46s & 0.1739 & 116m46s \\
     \midrule
    \multirow{2}{*}{Llama-3.1-70B\tnote{2} }& w/ static & 0.3449 & 103m8s & 0.1782 & 661m34s \\
    & w/o static & 0.1116 & 102m58s & 0.1689 & 661m26s \\
  \bottomrule
\end{tabular}
\begin{tablenotes}
\item[1,2] Both GPT-4o API calling and Llama-3.1-70B model weights downloading were executed as of this writing in October, 2024.
\end{tablenotes}
\end{threeparttable}
\end{table}

\subsection{Retrospective retrieval of health records}
The wide adoption of Electronic Health Record (EHR) greatly improves communication and availability of relevant information for both retrospective research and real-time clinical decision support. Accurate retrospective retrieval or search of EHRs is a fundamental task for evidence-based healthcare, clinical research, precision medicine, and community health management. For individuals, accurate access to rich and relevant case records enables real-time clinical decision support, sharing and analytics of appropriate information in a common way, contributing evidence for better treatment, and reducing costs for inappropriate and duplicative testing. For populations, storing and sharing health records in a common and accurate way facilitates early identification of emerging health issues, and hence agile response to evolving clinical practices, reducing costly or even deadly negligence and errors.

However, a simple use of conventional information retrieval (IR) techniques, such as in popular daily search engines, provides only limited benefits to clinical research and decision support. Low recall and imprecise results requires further heavy-duty data post-processing, therefore hinders the wide usage of retrospective retrieval of health records in real-world clinical setting. The major reason is that general-purpose IR is a language-level system. In a critical domain such as healthcare, we need a meaning-based retrieval system at the clinical domain level. One example from the field is as follows. Clinicians want to find ``historical patients with type 2 diabetes complicated by diabetic nephropathy''. A modern general-purpose IR system would likely word segment ``type 2'' apart from ``diabetes'', and assign non-trivial relevance scores to examples of ``type 1 diabetes'' (similar in dense retrieval in embedding space). From the clinical perspective, however, these two diseases have differences in causes, symptoms and treatment, thus ``type 1 diabetes'' results are imprecise. A standardized clinical terminology like our MedCT becomes a systematic approach to bringing semantic understanding from linguistic level to domain level. In our experiments, all clinical notes and health records with mentions of ``type 2 diabetes'' were accurately recognized and linked to the MedCT concept: ``\texttt{44054006 (find) $\mid$ Diabetes mellitus type II $\mid$ II \chinese{型糖尿病}}''.

In this experiment, we wish to validate and measure the value of the clinical terminology MedCT in the application of health record retrieval. A majority of retrospective retrieval of EHRs involves finding cases with similar or related diseases, in reference past evidences in testing, diagnosis, treatment and outcome. Therefore, from the MedCT-clinical-notes dataset (as described in Section~\ref{sec-truth}), we took a corpus of discharge summaries for this retrieval experiment. The corpus contains 13,863 examples or discharge notes, entered from all departments during the first quarter of year 2024 in a tertiary care hospital. The data was organized into relevant textual fields including bio, admission, treatment pathway, discharge summary and instruction. We interviewed a panel of 12 senior physicians to collected a set of 20 queries representative of real-world clinical practice and research. The clinical query set was chosen with non-trivial complexity such that a straightforward keyword match conceivably cannot yield satisfactory results. One example is ``post-stroke with pneumonia'', and the full query set can be found in Appendix~\ref{apdx-clinical-queries}.

We implemented two retrieval strategies, the classic sparse or dense retrieval and the MedCT-augmented retrieval. The sparse retrieval approach uses a bag-of-words retrieval function, predominantly BM25~\cite{Robertson:2009}, that ranks a corpus of documents based on the query terms appearing in each document. Our implementation is based on Elasticsearch~\cite{elastic:2024}, hence supports full text queries, including fuzzy, phrase or proximity matching. The dense retrieval integrates term-based search with semantic search in the embedding space, using a popular sentence transformers all-MiniLM-L6-v2~\cite{reimers:2019} for text embedding. Both the sparse and dense retrieval represents modern search technology. For the MedCT-augmented retrieval approach, we first offline tagged each document with MedCT concepts, and then indexed the list of concept ids along with texts per document. These annotated concepts should capture almost all relevant clinical information in the health records. At online retrieval time, we annotated full text queries with MedCT concepts, and then ranked documents with a hybrid strategy based on both text-based sparse or dense retrieval and strict concept id matching.

For evaluation, we asked the same panel clinicians to annotate relevant examples from a random sample of 2K discharge notes, for each of the 20 queries. This ground truth allows us to measure precision, recall and the balanced $F_1$ score as the performance metrics for our retrieval task, as reported in Table~\ref{tab-retrieval-metrics}. 

\begin{table}
\begin{threeparttable}
  \caption{EHR retrieval augmented with MedCT}
  \label{tab-retrieval-metrics}
  \begin{tabular}{lrrr}
    \toprule
    Retrieval method & Precision\tnote{1} & Recall\tnote{2} & $F_1$-score\tnote{3} \\
    \midrule
    Sparse & 0.5294 & 0.5015 & 0.5151 \\
    Dense & 0.0706 & 0.0995 & 0.0826 \\
    Hybird & 0.3882 & 0.2527 & 0.3061 \\
    MedCT-aug. & \textbf{0.6235} & \textbf{0.5745} & \textbf{0.5980} \\
  \bottomrule
\end{tabular}
\begin{tablenotes}
\item[1,2,3] All metrics are measured at top 10 retrieved results, representative of a typical search scenario.
\end{tablenotes}
\end{threeparttable}
\end{table}

Our experiments show that retrieval augmented with MedCT graph substantially outperforms modern text-based search. In particular, MedCT boosts the search recall by a 15\% lift over sparse retrieval. Clinical information retrieval is a challenging task, since population positive ratio is by nature very low, or disease is rare event. For our testing query set, the positive ratio is typically below 1\%. For a ``needle-in-a-haystack'' task like in our setting, recall is more critical than precision as a performance measure. This is also true for the real-world clinical decision making and research. We also found empirically that dense retrieval worked poorly alone, neither helped when integrated with sparse method. We hypothesized two causes as follows. The embedding model is still quite general, therefore not specialized enough to capture the semantics in clinical texts. Also, clinical documents such as EHRs are not typical natural language texts, e.g., normally contain sufficient special formatting and jargons, mostly proprietary, not available from web for training models. These subtleties are often seen in high-resourced domains such as healthcare, that warrants the development of a truth model for various fundamental tasks. We further show some visual examples of different retrieval methods as in Figure~\ref{fig-ehr-retrieval-snippet}.

\begin{figure}
\begin{mdframed}
 \footnotesize
\textbf{Query: \chinese{脑梗死后合并肺部感染} (Post-Stroke with Pneumonia)} \\
\\
\textbf{Sparse retrieval results: }\\
\begin{enumerate}
\item note_id: 10419-51\quad discharge_diagnosis: \chinese{1、右侧输尿管结石伴有积水和\colorbox{black!15}{感染} 2、脂肪肝 3、双肾囊肿 4、陈旧性脑出血 5、急性\colorbox{black!15}{脑梗}塞}\ldots (score: 18.67)\\
\item note_id: 6338792-3\quad discharge_diagnosis: \chinese{1.急性后循环\colorbox{black!15}{脑梗死}，基底动脉闭塞 右侧侧脑室旁、两侧小脑半球、脑干\colorbox{black!15}{脑梗死} 2.高血压病 3.脑积水}\ldots (score: 17.36)\\
\item note_id: 1077041-4\quad discharge_diagnosis: \chinese{1、构音障碍 2.乙肝表面抗原阳性 3.脂肪肝 、肝功能异常。患者头颅MR未见新发\colorbox{black!15}{梗死}，且患者目前仍感口齿含糊，既往劳累时也有这种情况，目前暂不考虑新发\colorbox{black!15}{梗死}，予停用相关药物；其余予护肝治疗。}\ldots (score: 17.04)\\
\end{enumerate}

\textbf{MedCT-aug. query: \colorbox{red!30}{\chinese{脑梗死}}[432504007]\chinese{后合并}\colorbox{blue!30}{\chinese{肺部感染}}[128601007]} \\
\\
\textbf{MedCT-aug. retrieval results:} \\
\begin{enumerate}
\item note_id: 259694-136\quad discharge_diagnosis: \chinese{1.脓毒症 感染性休克 \colorbox{blue!30}{肺部感染} 肺水肿 胸腔积液  2.腹膜透析相关腹膜炎 腹膜透析 3.慢性肾脏病5期 肾性贫血 维持性腹膜透析状态 肾性骨病 心功能不全 4.冠状动脉粥样硬化性心脏病 KillipI级 心肌梗死 5.胆囊结石 脂肪肝 6.亚临床甲状腺功能减退  7.高血压 8.甲状腺结节 9.糖尿病 糖尿病伴心脏并发症 糖尿病性肾病 糖尿病性周围神经病 10.心律失常：心动过缓 房室传导阻滞 11.癫痫 12.梅毒 13.\colorbox{red!30}{脑梗}个人史 脑缺血灶 14.肺结节 15.胆囊结石 胆囊炎}\ldots (score: 24.05)\\
\item note_id: 6339349-3\quad discharge_diagnosis: \chinese{1.左侧椎动脉闭塞\colorbox{red!30}{脑梗死} 左侧小脑半球脑梗死伴出血转化 脑干梗死 脑积水 急性呼吸衰竭 2.右侧椎动脉远段、基底动脉狭窄，右侧大脑前动脉A2段管腔中重度狭窄考虑 3.高血压 4.\colorbox{blue!30}{肺部感染} 5.高钠血症 6.肝功能不全}\ldots (score: 23.91)\\
\item note_id: 725323-11\quad discharge_diagnosis: \chinese{1.高钠血症 2.\colorbox{blue!30}{肺部感染} 3.肾功能不全 4.肝功能不全 5.高血压 6.\colorbox{red!30}{脑梗死}后遗症 7.颅脑外伤术后 8.胆囊结石 胆囊炎 9.低蛋白血症}\ldots (score: 23.88)\\
\end{enumerate}
\end{mdframed}
\caption{A snippet from sparse and MedCT-aug retrievals}
\label{fig-ehr-retrieval-snippet}
\end{figure}

As visualized in the above example, the search query essentially was to retrieve patient examples with two clinical findings simultaneously: ``cerebral infarction'' (concept id: 432504007) and ``pulmonary infection'' (concept id: 128601007). However, none of the top three results from the sparse retrieval method is true positive, containing only partial or none of the searching concepts. Also as the grey color-coded terms show, these false positives are due to matching of terms, often multiple times, at the natural language level, not the domain concept level. Many matched terms are too general, such as ``infection", while some matches appear in negative mentions, e.g., ``not considered cerebral infarction". On the other hand, the MedCT-augmented retrieval accurately recalls true positives at all top three results, by matching concept ids between query and documents, as color-coded in the above example in Figure~\ref{fig-ehr-retrieval-snippet}.


\subsection{Health records auto-generation by LLMs}
Next we evaluate MedCT on the task of health records auto-generation by LLMs. There have been emerging applications of LLMs in the healthcare domain~\cite{Clusmann:2023}, from clinical workflow (e.g., clinical notes transcribing and health records generation), patient care (e.g., triage and follow-up by AI bot), to medical research (e.g., data retrieval and analytics). We consider the task of health records generation. Nearly half of physician’s time is devoted to digital paperwork, rather than direct patient care~\cite{Sinsky:2016aa}. The statistics is even worse in regions and countries with shortfall of health workforce. In our field study at a tertiary hospital in a rural area in China, reportedly near 90\% of residency doctors' time is absorbed in writing clinical notes and medical records. Among various health records, discharge summary is arguably the most important document a hospitalist writes. The discharge summary is a semi-structured narrative document for communicating clinical information about patients in the hospital. However, nowadays hospitalists have little time to write good quality discharge summaries, along with often delay to deliver to downstream outpatient physicians, causing disruption in the continuity of care and risking poor patient outcomes.

Therefore it is appealing to apply LLMs to generate health record drafts for physicians, to review, mild edit and submit. This is a text summarization task with moderate complexity yet high practical significance~\cite{Chen2024aa}. A good model, likely with domain or task-specific training, would both speed clinical documentation and improve the quality of health records. In our deployment to a tertiary care hospital in Zhejiang, China, hospitalists reported about 40\% reduction in time spent in writing discharge summaries with the help of LLM generation, while observing improvement in both quality and information density.

However, general-purpose LLMs, if used as-is in a vanilla fashion, typically cannot meet the safety requirements of the medical domain~\cite{Stanceski:2024,Singhal:2023aa,Raza:2024}. In our controlled experiments, for instance, we observed that vanilla LLMs hallucinated medical misinformation such as made-up procedures (e.g., Laparoscopy for radical pulmonary surgery) and medications (e.g., Cefuroxime) in discharge summary generation. This represents a significant risk of applying LLMs in an ignorant way in mission-critical domains like healthcare. Many of these hallucinations were trivial to identified by qualified physicians, which also symbolizes the large gap between human intelligence and LLMs, especially in domain knowledge.

In order to address the hallucination problem intrinsic to LLMs, we guide the LLM generation with a knowledge graph as source of truth. We believe that our approach brings together the strengths of both worlds, LLMs and specialized small models; and moreover presents a systematic and measurable way to minimize hallucination. We first pretrained a LLM, namely Tigerbot-3~\cite{chen:2024ab}, continually from Llama-3.1~\cite{llama3:2024} to strengthen biomedical base knowledge (with medical training data in Table~\ref{tab-medbert-data}) and multilingual coverage (especially Chinese for our applications). We then fine-tuned an instruction-tuned generative model for both general-purpose (e.g., question-answering, chat and summarization) and domain-specific clinical tasks (including discharge summary auto-generation). For the task of discharge summary generation, we prompt the instruct model with the input context consisting of detailed hospital course and discharge diagnosis, and instruct the model to generate brief hospital course. The brief hospital course is the major section that needs to be summarized from the lengthy detailed records, and typically is also the most time-consuming part. We conducted a controlled experiment to compare a vanilla zero-shot LLM prompting and a MedCT-guided generation. As illustrated in the detailed prompts in Figure~\ref{fig-notegen-prompts}, the MedCT-guided generation instructs the LLM to attend to major clinical concepts such as ``chief complaint'' and ``physical examination'' and therefore should capture key clinical information more comprehensively and accurately.

\begin{figure}
\centering
\begin{mdframed}
\small
\begin{verbatim}
# Zero-shot prompt for vanilla LLM approach:
{input context: hospital course, discharge diagnosis}

The above is a detailed hospital course and discharge 
diagnosis from a medical record. Please summarize it 
into an accurate and concise medical summary. The 
summary should include the reason for admission, 
basis for diagnosis, main treatment measures and their 
effects, changes in condition, and status at discharge.

Medical summary:

# Zero-shot prompt for MedCT-guided LLM approach:
{input context: hospital course, discharge diagnosis}

The above is a detailed hospital course and discharge 
diagnosis from a medical record. Please summarize it 
into an accurate and concise medical summary. The 
summary should include the reason for admission, 
basis for diagnosis, main treatment measures and their 
effects, changes in condition, and status at discharge.

Also, the above medical records include the following 
entities, please include these medical entities in the 
medical course summary.
{entity mentions from the input}

Medical summary:
\end{verbatim}
\end{mdframed}
\caption{Prompts for discharge summary auto-generation}
\label{fig-notegen-prompts}
\end{figure}

To evaluate the generation results, we recruited a panel of nine hospitalists to review summary generations from the above two methods, along with the human summary by doctors, in a blind fashion, and then cast Likert scale to each testing example. Evaluating text summarization models is nontrivial, especially with automatic metrics. In general domains, such as TL;DR Reddits and CNN/DM news article summarization, previous works have used ROUGE or reward models to predict human preference~\cite{Stiennon:2020aa,Wu:2021aa}. But these metrics are only rough proxies to real human perceived summary quality, and should not apply indiscriminately to different domains or even different tasks. For example, in book and news article summarization, coherence is often used to measure how easy the summary is to read on its own. But for the task of health record summarization in the clinical domain, with the time pressure and norm use of medical abbreviation and terminology, conciseness and clarity weigh more than coherence. After three sessions of panel discussions, we developed a set of metrics for our health record summarization task. The metrics cover both general language quality and clinical significance, from perspectives of accuracy, completeness, clarity, relevance, conciseness, and clinical depth. Moreover, our evaluation metrics, along with the annotated preference dataset, shall be instrumental to develop automatic metrics for text summarization in the clinical domain. A detailed guideline for labeler rating can be seen in Appendix~\ref{apdx-ehrsum-rating}. 

Our evaluation dataset contains 91 examples of discharge notes, with detailed hospital courses and discharge diagnosis as raw input (denoted as raw), along with discharge summaries written by human hospitalists (denoted as human). We then infer the underlying LLM with two prompting approaches as in Figure~\ref{fig-notegen-prompts} (denoted as LLM and MedCT, respectively). The average length in character is 3,545  for the input clinical notes, 542 for the human summary, 274 for the vanilla LLM, and 317 for the MedCT method. LLMs tend to condense more than human, while the MedCT-augmented method conveys richer information than simple LLM prompting. This is as expected, since we instruct the LLM to preserve information regarding clinical entities. For a rapid computerizable evaluation, we compute cosine similarities, in the embedding space, between input, human and machine generations. With similar compression rates, cosine similarity is a reasonable proxy to how well a summary captures the original text's main points. as shown in Table~\ref{tab-notegen-cosine}.

\begin{table}
 \caption{EHR summarization cosine similarity}\label{tab-notegen-cosine}
 \small
\begin{tabular}{lrrrrr}
\toprule
\multirow{2}{*}{GPT-4o} & \multicolumn{3}{c}{\textbf{Raw}} & \multicolumn{2}{c}{\textbf{Human}} \\            
\cmidrule(r){2-4}\cmidrule(r){5-6}
 & Human & LLM & MedCT & LLM & MedCT \\
 \midrule
MedBERT & 0.8940 & 0.8984 & \textbf{0.9288} & 0.9242 & \textbf{0.9257} \\
all-MiniLM & 0.6846 & 0.6947 & \textbf{0.8247} & \textbf{0.7618} & 0.7140 \\
\midrule
 \multirow{2}{*}{Llama-3.1-70B} & \multicolumn{3}{c}{\textbf{Raw}} & \multicolumn{2}{c}{\textbf{Human}} \\            
\cmidrule(r){2-4}\cmidrule(r){5-6}
 & Human & LLM & MedCT & LLM & MedCT \\
 \midrule
MedBERT & 0.8940 & 0.8897 & \textbf{0.9066} & \textbf{0.9163} & 0.9150 \\
all-MiniLM & 0.6846 & 0.6799 & \textbf{0.7506} & \textbf{0.7521} & 0.7385 \\
 \midrule
 \multirow{2}{*}{Tigerbot-3-70B} & \multicolumn{3}{c}{\textbf{Raw}} & \multicolumn{2}{c}{\textbf{Human}} \\            
\cmidrule(r){2-4}\cmidrule(r){5-6}
 & Human & LLM & MedCT & LLM & MedCT \\
 \midrule
MedBERT & 0.8940 & 0.9013 & \textbf{0.9147} & \textbf{0.9156} & 0.9104 \\
all-MiniLM & 0.6846 & 0.6576 & \textbf{0.7388} & \textbf{0.6837} & 0.6836 \\
\bottomrule
\end{tabular}
 \end{table}
 
We experimented with three LLMs, namely GPT-4o, Llama-3.1-70B and Tigerbot-3-70B. These three choices represent proprietary, open-source, and domain-specialized model families, respectively. We also used two SOTA embedding models, the general-purpose all-MiniLM-L6-v2 and our specialized trained MedBERT (see Section~\ref{sec-medbert}). With respect to raw health records, our MedCT-augmented generation achieves best cosine scores, notably even higher than human summaries. This finding is particularly encouraging in that graph-augmented LLMs may reach human-like intelligence in clinical text summarization task. But human summarization from the field in real-world clinical setting should not be deemed as gold standard or ground truth, for various realistic reasons such as time pressure and variance in experience. Also note that, the specialized biomedical model MedBERT yields higher similarity scores than the general-purpose model. This is because domain-specialized training casts more attention to clinical semantics, via both vocabulary and model weights. Furthermore once again, for tasks that require deep domain knowledge such as clinical notes summarization, proprietary and open-source models empirically yielded comparable performance, if we use $cosine(\text{human},\text{LLM})$ as a proxy. Meanwhile, with moderate domain specialization, our cost-effective alternative (Tigerbot-3-70B) achieved quite competitive results with best $cosine_\text{MedBERT}(\text{raw},\text{LLM})$.

\begin{figure}
\centering
\subfigure[LLM vs. Human win-tie-loss rate]{\includegraphics[width=0.48\linewidth]{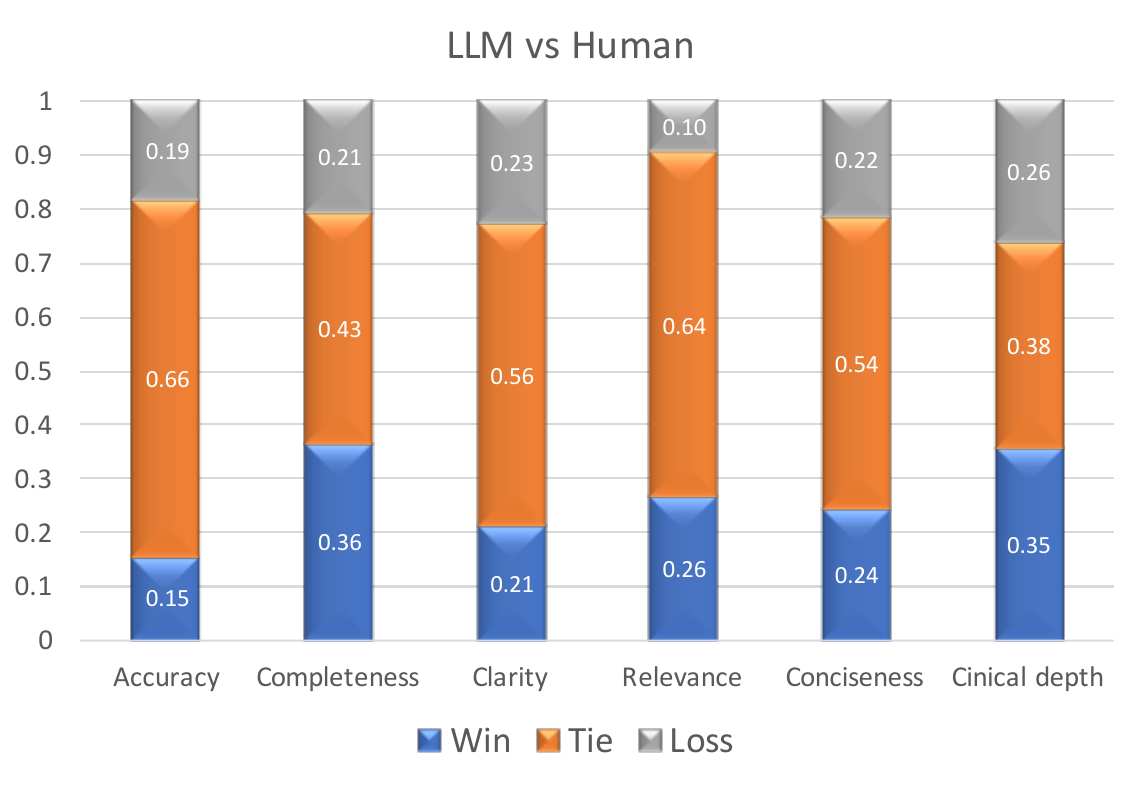}
\label{fig-notegen-winrate-a2h}}
\subfigure[MedCT vs. Human win-tie-loss rate]{\includegraphics[width=0.48\linewidth]{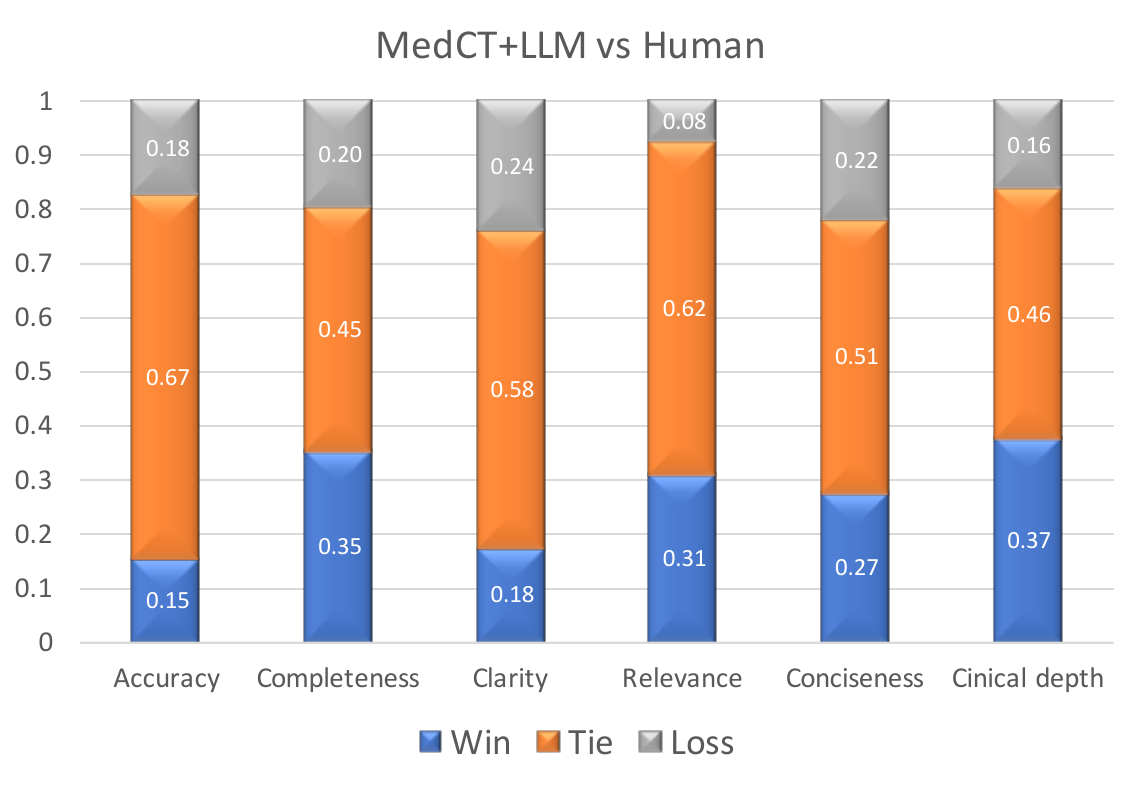}
\label{fig-notegen-winrate-b2h}}
\caption{EHR generation win-tie-loss rates over human}
\label{fig-notegen-winrate}
\end{figure}

While the programable cosine similarity gives a rapid proxy to summarization quality, especially for model iteration and comparison, the gold standard is still human review. We distributed the 91 testing examples to nine hospitalists with tenure from ten years or above. For each example, we gave original input clinical notes, and three summary generations, from human doctors, simple LLM (Tigerbot-3), and LLM augmented with MedCT graph (denoted as human, LLM and MedCT, respectively). More importantly, the review was conducted in a blind manner. The three generations were randomly shuffled per example (only organizers held the true orders), and hence reviewers did not know which model or human generated the summary to be scored. The physician reviewers were instructed to rate summaries using 5-point Likert-scale by real-world clinical standards in reference to the guideline illustrated in Appendix~\ref{apdx-ehrsum-rating}. One of the entries is factually human summary by real doctor anyway. The results are shown in Table~\ref{tab-notegen-scores}. To normalize the scores over examples, we also computed win-tie-loss rates over human summary, as plotted in Figure~\ref{fig-notegen-winrate}. Further, we show an example of input and generation snippets, with a focus on treatment outcomes, in Figure~\ref{fig-notegen-exp}.

\begin{table}
 \caption{EHR summarization human review scores}\label{tab-notegen-scores}
\begin{tabular}{lrrr}
\toprule
& Human & LLM & MedCT \\
\midrule
Accuracy	& 4.42 & 4.41 & \textbf{4.42} \\
Completeness	& 3.99 & 4.19 & \textbf{4.20} \\
Clarity & 4.44 & \textbf{4.45} & 4.40 \\ 
Relevance & 4.56 & 4.77 & \textbf{4.84} \\
Conciseness & 4.23 & 4.27 & \textbf{4.33} \\
Cinical depth & 3.95 & 4.07 & \textbf{4.19} \\
\midrule
\textbf{Overall} & 25.58 & 26.15 & \textbf{26.36} \\
\bottomrule
\end{tabular}
 \end{table}
 
\begin{figure}
\centering
\begin{mdframed}
\small
\textbf{Input clinical notes:} \\
Patient_id: 13 \\
Hospital course 1: 2024/1/11 10:48 \\
Chief complaint: The patient's temperature was the highest yesterday at 38°C, and his stool was slightly better than before.\\
Diagnosis: 1. Open craniocerebral injury, \ldots 3. Contusion of both lungs, bilateral pleural effusion with incomplete expansion of both lower lungs, and \colorbox{red!30}{pulmonary infection}\ldots\\
\\
\textbf{Human summary:} \\
After admission, he was given mannitol to lower intracranial pressure,\ldots insulin combined with metformin to lower blood sugar, and parenteral nutrition support.\\
\\
\textbf{LLM summary:} \\
In terms of changes in condition, although the patient suffered from various complications, after careful treatment by the doctor, the patient gradually regained consciousness and was able to perform some basic self-care. Finally, the patient was discharged from the hospital on January 28, 2024, and planned to continue rehabilitation training after discharge.\\
\\
\textbf{LLM+MedCT summary:} \\
After treatment, the patient's condition gradually stabilized, but there were still some problems, such as repeated fever, low level of consciousness, and \colorbox{red!30}{lung infection}. Finally, the patient was discharged from the hospital on January 28, 2024, and planned to undergo rehabilitation training after discharge.
\end{mdframed}
\caption{A snippet from clinical notes summarization}
\label{fig-notegen-exp}
\end{figure}

Overall, our MedCT-guided LLM approach achieves highest human ratings, winning five out of six review dimensions. Notably, the gains from the perspectives of ``clinical depth'' and ``relevance''  are particularly substantial, over both LLM and human generations. By additionally prompting LLM with MedCT-recognized clinical entities, the underlying models exhibited precise attentions to clinical concepts in the free-text notes. As illustrated in the snippets in Figure~\ref{fig-notegen-exp}, the MedCT-augmented approach captured the finding of ``pulmonary infection'' in hospital course summary, while simple LLM and human missed this concept. From the input notes of this patient, ``pulmonary infection'' is one of the complications through most of her or his hospital course. Both machine generations yielded comparable or better summarization with respect to factual human summaries. The win-or-tie rate of LLM approach is 0.59 over human, and even higher augmented by MedCT with a good odd of 0.68 being at least as good as human. This finding confirms with previous computerized evaluation of cosine similarity, suggesting that LLMs with domain specialization indeed can help clinical workflow tasks such as health record auto-generation.

\section{Conclusions}
Healthcare arguably remains one of the most prosperous domains beneficial from the rapid development of artificial intelligence (AI) in general and large language models (LLMs) in particular. 

\epigraph{``We're going to have a family doctor who's seen a hundred million patients and they're going to be a much better family doctor.''~\cite{Hinton:2023}}{\textit{Geoffrey Hinton}}

However, even the state-of-the-art general foundation models, e.g., GPT-4o and Llama-3.1, merely scratch the surface encoding deep domain knowledge from high-resourced yet largely private domains such as healthcare. Moreover, the probabilistic root of LLMs, hence the tendency of hallucination, hinders their wide practical adoption in privacy and safety critical tasks. In the context of leveraging LLMs' extraordinary capabilities of semantic understanding, generativeness and interactiveness, while ensuring their safety, unbiasedness, and honesty in real-world applications, we developed and released MedCT. To the best of our knowledge. MedCT is the world's first clinical terminology built for and grounded from non-English community, specifically Chinese. We presented our comprehensive approach to building the clinical terminology knowledge graph, truth grounding and optimizing from real clinical data, training models for named entity recognition and linking to the graph. Our approach leverages LLMs as an integral part of development tools, along with abundant real-world clinical data and annotations by experienced physicians. Consequently, our MedCT models achieve new state-of-the-art in medical NER and NEL tasks (w.r.t. BiomedBERT and SciBERT etc.), especially in a rapid and cost-efficient manner (w.r.t. SNOMED CT etc.). 

The values of our MedCT clinical terminology are even more pronounced in complementing LLM applications in clinical setting. A knowledge graph such as MedCT not only injects clinical domain commonsense into foundation models, but also as a source of truth gouges model generations to be more safe, truthful and reliable. We deployed MedCT in a wide variety of clinical applications, including clinical information retrieval and document summarization. Our findings from human blind reviews are inspiring, in that MedCT-augmented LLMs can achieve human-like or even better results in various tasks in clinical workflow and research tasks. Meanwhile, we also found that general-purpose LLM is no silver bullet, especially for domains with knowledge depth. As our approach stands, practical yet mission-critical applications of LLM still require domain specialization, for example in conjunction with classical yet surgical machine learning techniques.

We believe that we are at the dawn of unleashing the values of AI and LLMs for great humanity. In the hope of facilitating further development in the healthcare domain, we open-source release the MedCT suite of models and datasets~\footnote{For a detailed description and ongoing releases, please see MedCT repository:\\Github: \url{https://github.com/TigerResearch/MedCT};\\Huggingface: \url{https://huggingface.co/collections/TigerResearch/medct-6744641d6f19b9d70a56f848}}, which include:
\begin{enumerate}
\item The MedCT bilingual (English and Chinese) clinical terminology dictionary, with 223K medical concepts.
\item The MedCT named entity recognition (NER) model: MedLink.
\item A biomedical foundation model: MedBERT, that achieves state-of-the-art performance in a variety of downstream tasks, e.g., clinical NER/NEL, search, and summarization.
\item Our MedCT-clinical-notes dataset, including: 
	\begin{itemize}
	\item For the NER and NEL tasks, 7.4K real-world clinical notes in Chinese, and 61K entity mention annotations per MedCT graph.
	\item For the search task, 20 clinical queries, and 2K discharge notes with relevance annotations.
	\item For the clinical note summarization task, 91 raw discharge notes and summaries by human, LLM and MedCT-augmented generations, along with preference Likert-scale annotated by human physicians.
	\end{itemize}
\end{enumerate}

\bibliographystyle{ACM-Reference-Format}
\bibliography{medct-sigconf}


\appendix

  \section{Examples of Chinese Electronic Health Records (EHR)}
  \label{apdx-ehr}
  \begin{mdframed}
   \footnotesize
  \subsection*{\chinese{入院记录：}(Admission record:)}
  \chinese{性别 (Gender)：女}\quad \chinese{职业 (Occupation)：自由职业者}\quad \chinese{出生日期 (Date of birth)：1996/10/1}\quad \chinese{婚姻状况 (Marital status)：已婚}\\
  \chinese{出生地 (Place of birth)：浙江省金华市义乌市}\quad \chinese{民族 (Ethnicity)：汉族}\quad \chinese{目前使用药物 (Medications on admission)：无}
    
  \subsection*{\chinese{主诉：}(Chief complaint:)}
  \chinese{右侧胸痛6小时余}
  
 \subsection*{\chinese{现病史：}(History of present illness:)}
 \chinese{患者6小时之前无明显诱因下出现右侧胸刺痛，无明显咳嗽咳痰，无发热寒战，无乏力纳差，无头晕头痛，无恶心呕吐，休息后无明显缓解，为进一步治疗来我院就诊，拟“胸痛”收住入院。完善相关检查，肺部CT提示：右侧气胸，肺被压缩约为70\%。右肺渗出改变。于急诊行吸氧、补液等治疗，密切监护。病情平稳后转入我科拟行进一步诊治。
病程中，未进食，未睡眠，二便无殊，近期体重无明显增减。}

\subsection*{\chinese{既往史：}(Past medical history:)}
\chinese{既往自发性气胸史半年余，否认“高血压”、“糖尿病”、“心脏病”、“冠心病”、“脑血管意外”、“慢性支气管炎”、“肾病”等病史}

\subsection*{\chinese{个人史：}(Social history:)}
\chinese{出生于江西省上饶市弋阳县，生长于江西省上饶市弋阳县，否认异地长期居留史，文化程度高中，职业其他，否认吸烟史、否认饮酒史、否认疫区居留史、否认疫水、疫源接触史、否认其他特殊嗜好否认不洁性交史、否认长期放射性物质、毒物接触史、否认粉尘吸入史。}

\subsection*{\chinese{家族史：}(Family history:)}
\chinese{父母健在，1姐健在，体健，否认类似疾病史，否认家族中Ⅱ系Ⅲ代传染病、遗传病、精神病、家族性疾病及肿瘤性疾病史。}

\subsection*{\chinese{初步诊断：}(Admission diagnosis:)}
\chinese{气胸}

\subsection*{\chinese{实验室检验和辅助检查：}(Laboratory tests and auxiliary examinations:)}
\chinese{（2024-01-01 11:08，本院）行胸部CT平扫＋三维重建检查提示：右侧气胸，肺被压缩约为70\%。右肺渗出改变。}\\
\chinese{（2024-01-01 11:30，本院）行（急诊）常规十二导心电图检测检查提示：1.窦性心律； ；2.正常心电图。}

\textbf{\chinese{胸部正位_诊断影像：}}
\begin{itemize}
	\item \chinese{右侧气胸引流后复查，建议治疗后随诊复查。}
	\item \chinese{右中肺野扁状致密影，外物投影可能。}
\end{itemize}

\subsection*{\chinese{诊疗计划：}(Diagnosis and treatment plan:)}
\begin{enumerate}
\item \chinese{检查计划：完善血常规、生化、凝血功能等检查}
\item \chinese{治疗计划：VTE低危，予一般预防；拟行胸腔穿刺闭式引流，若病情无好转或进一步进展考虑手术治疗。}
\item \chinese{预期治疗结果：缓解症状，气胸吸收。}
\item \chinese{预期住院天数：4-5天}
\item \chinese{预期费用：5000-10000元}
\item \chinese{转诊或出院计划：病情恢复稳定后出院}
\end{enumerate}

\ldots \ \ \ldots

\subsection*{\chinese{住院经过：}(Brief Hospital Course:)}
\chinese{入院后完善相关检查，排外手术禁忌后于2024.1.1于我院行胸腔穿刺闭式引流，患者胸闷明显改善，拒绝手术治疗后予带药出院。}

\subsection*{\chinese{出院诊断：}(Discharge Diagnosis:)}
\begin{enumerate}
\item \chinese{自发性气胸}
\item \chinese{肺大疱破裂}
\end{enumerate}

\subsection*{\chinese{出院情况：}(Discharge Conditions:)}
\chinese{神志清，精神可，双肺呼吸音清，无明显干湿音。}

\subsection*{\chinese{出院医嘱：}(Discharge Instructions:)}
\chinese{萘丁美酮胶囊  1克  口服  每日一次 1盒；}

\subsection*{\chinese{随访计划：}(Follow-up Instructions:)}
\chinese{出院2周门诊复诊复查胸片，如有发热、剧烈胸闷胸痛等不适，及时就诊。}
\end{mdframed}

\section{Clinical queries for EHR retrieval}\label{apdx-clinical-queries}
\begin{table}[H]
  \caption{Queries for health record retrieval}
  \label{tab-clinical-queries}
  \tiny
  \begin{tabular}{rl}
    \toprule
    No. & Query (Chinese English) \\
    \midrule 
    \multirow{2}{*}{1} & \chinese{2型糖尿病并发糖尿病肾病} \\
    & Type 2 Diabetes Mellitus with Diabetic Nephropathy \\
    \midrule 
    \multirow{2}{*}{2} & \chinese{慢性阻塞性肺疾病急性加重期伴呼吸衰竭} \\
    & Chronic Obstructive Pulmonary Disease with Acute Exacerbation and Respiratory Failure \\
     \midrule 
     \multirow{2}{*}{3} &  \chinese{高血压伴左心室增厚} \\
     & Hypertension with Left Ventricular Hypertrophy \\
    \midrule 
    \multirow{2}{*}{4} & \chinese{肺癌伴脑继发性恶性肿瘤} \\
    & Lung Cancer with Brain Metastasis \\
     \midrule 
     \multirow{2}{*}{5} &  \chinese{结直肠癌术后出现肝脏继发恶性肿瘤} \\
     & Colorectal Cancer Postoperative with Liver Metastasis \\
   \midrule 
    \multirow{2}{*}{6} & \chinese{系统性红斑狼疮伴狼疮性肾炎} \\
    & Systemic Lupus Erythematosus with Lupus Nephritis \\
    \midrule 
    \multirow{2}{*}{7} & \chinese{慢性乙型病毒性肝炎及肝硬化并发食管胃底静脉曲张} \\
    & Chronic Hepatitis B and Cirrhosis with Esophageal and Gastric Varices \\
    \midrule 
    \multirow{2}{*}{8} & \chinese{胰腺炎合并高脂血症} \\
    & Pancreatitis with Hyperlipidemia \\
    \midrule 
    \multirow{2}{*}{9} & \chinese{脑梗死后合并肺部感染} \\
    & Post-Stroke with Pneumonia \\
    \midrule 
    \multirow{2}{*}{10} & \chinese{帕金森病合并老年痴呆} \\
    & Parkinson's Disease with Dementia \\
    \midrule 
    \multirow{2}{*}{11} & \chinese{冠状动脉粥样硬化性心脏病伴心房颤动} \\
    & Coronary Artery Disease with Atrial Fibrillation \\
    \midrule 
    \multirow{2}{*}{12} & \chinese{妊娠期高血压并发HELLP综合征} \\
    & Gestational Hypertension with HELLP Syndrome \\
    \midrule 
    \multirow{2}{*}{13} & \chinese{急性心肌梗死行经皮腔内冠状动脉成形术（PTCA）} \\
    & Acute Myocardial Infarction with Percutaneous Transluminal Coronary Angioplasty (PTCA) \\
    \midrule 
    \multirow{2}{*}{14} & \chinese{消化道出血并发失代偿性休克} \\
    & Gastrointestinal Bleeding with Decompensated Shock \\
    \midrule 
    \multirow{2}{*}{15} & \chinese{甲状腺乳头状癌行甲状腺根治术} \\
    & Papillary Thyroid Carcinoma with Thyroidectomy \\
    \midrule 
    \multirow{2}{*}{16} & \chinese{乳腺恶性肿瘤术后化学治疗} \\
    & Postoperative Chemotherapy for Breast Cancer \\
    \midrule 
    \multirow{2}{*}{17} & \chinese{重症肺炎伴呼吸衰竭} \\
    & Severe Pneumonia with Respiratory Failure \\
    \midrule 
    \multirow{2}{*}{18} & \chinese{髋骨骨折手术后并发下肢静脉血栓形成} \\
    & Postoperative Hip Fracture with Deep Vein Thrombosis \\
    \midrule 
    \multirow{2}{*}{19} & \chinese{急性髓系白血病并发肺曲霉菌感染} \\
    & Acute Myeloid Leukemia with Pulmonary Aspergillosis \\
    \midrule 
    \multirow{2}{*}{20} & \chinese{输尿管结石伴有积水和感染} \\ 
   & Ureteral Calculi with Hydronephrosis and Infection \\
    \bottomrule
\end{tabular}
\end{table}

  
  \section{A guideline for rating hospital course summary}
  \label{apdx-ehrsum-rating}
  \begin{mdframed}
  \footnotesize
  \subsection*{Accuracy}
  \begin{enumerate}
  \item The summary contains \textbf{many} errors or misleading informations.
  \item The summary contains \textbf{some} errors or misleading informations.
  \item The summary is \textbf{almost accuracy} and contains \textbf{a few} errors or misleading informations.
  \item The summary is \textbf{accurate} and contains \textbf{very few} errors or misleading informations.
  \item The summary is \textbf{completely accurate} and contains \textbf{no errors} or misleading information.
  \end{enumerate}
  
  \subsection*{Completeness}
   \begin{enumerate}
  \item The summary \textbf{misses most} of the important clinical details.
  \item The summary \textbf{misses some} important clinical details.
  \item The summary \textbf{contains most} of the important details, but a few are missing.
  \item The summary \textbf{contains almost all} important details with only a few omissions.
  \item The summary \textbf{contains all} important clinical details and nothing is missed.
   \end{enumerate}
   
  \subsection*{Clarity}
  \begin{enumerate}
  \item The summary is \textbf{difficult to understand}, the language is unclear, and terminology is used inappropriately.
  \item The summary is \textbf{a bit difficult to understand}, the language is not clear enough, and the terminology is used incorrectly.
  \item The summary is \textbf{basically clear}, but some places are not clear enough.
  \item The summary is \textbf{clear and understandable}, the language is fluent, and the terminology is used appropriately.
  \item The summary is \textbf{very clear}, the language is concise and the terminology is precise.
  \end{enumerate}
   
   \subsection*{Relevance}
   \begin{enumerate}
  \item The summary is \textbf{irrelevant} to clinical decision-making and lacks relevant information.
  \item The summary \textbf{has little relevance} to clinical decision-making and there is insufficient relevant information.
  \item The summary is \textbf{relevant} to clinical decision-making, but the relevant information is not comprehensive enough.
  \item The summary is \textbf{closely related to} clinical decision-making and contains most of the relevant information.
  \item The summary is \textbf{highly relevant to} clinical decision-making and contains all necessary relevant information.
   \end{enumerate}
   
   \subsection*{Conciseness}
   \begin{enumerate}
  \item The summary is lengthy and contains \textbf{a lot of unnecessary} information.
  \item The summary is lengthy and contains \textbf{some unnecessary} information.
  \item The summary is basically concise, but contains a \textbf{small amount of redundant} information.
  \item The summary is concise, with a moderate amount of information and \textbf{very little redundant} information.
  \item The summary is very concise, with just the right amount of information and \textbf{no redundant information}.
   \end{enumerate}
   
   \subsection*{Clinical depth}
   \begin{enumerate}
  \item \textbf{No diagnostic analysis} or treatment options are reflected in the summary.
  \item There is \textbf{a simple analysis} or plan in the summary, but it lacks details.
  \item There are \textbf{certain analyzes} and plans in the summary, involving some clinical reasoning.
  \item The summary has \textbf{detailed analysis} and protocols demonstrating sound clinical reasoning.
  \item The summary features \textbf{comprehensive analysis} and scenarios that demonstrate deep clinical reasoning and insight.
   \end{enumerate}
\end{mdframed}

\end{document}